\documentclass{egpubl}
\usepackage{sca2026}

\CGFccby
\usepackage[T1]{fontenc}
\usepackage{dfadobe}  
\usepackage{amssymb}
\usepackage{booktabs}
\usepackage{cite}  % comment out for biblatex with backend=biber
\usepackage{graphicx}
\usepackage{amsmath}
\BibtexOrBiblatex
\electronicVersion
\PrintedOrElectronic
\usepackage{egweblnk} 
\makeatletter
\providecommand{\p@EGlocalpagenumber}{}
\def\ps@titlepage{%
  \let\@oddhead\@empty
  \let\@evenhead\@empty
  \def\@oddfoot{\hfil\thepage\hfil}
  \let\@evenfoot\@oddfoot
}

\let\oldps@headings\ps@headings
\def\ps@headings{%
  \oldps@headings
  \def\@oddfoot{\hfil\thepage\hfil}
  \def\@evenfoot{\hfil\thepage\hfil}
}
\makeatother

\title[Neural Assistive Impulses: Synthesizing Exaggerated Motions for
Physics-based Characters]%
      {Neural Assistive Impulses: Synthesizing Exaggerated Motions for
Physics-based Characters}

\author[Z. Wang \& B. Benes]
{\parbox{\textwidth}{\centering Zhiquan Wang, Bedrich Benes$^{1}$
        }
        \\
{\parbox{\textwidth}{\centering $^1$ Purdue University\\
       }
}
}
\begin{document}

% uncomment for using teaser
\teaser{
 \includegraphics[width=0.99\linewidth]{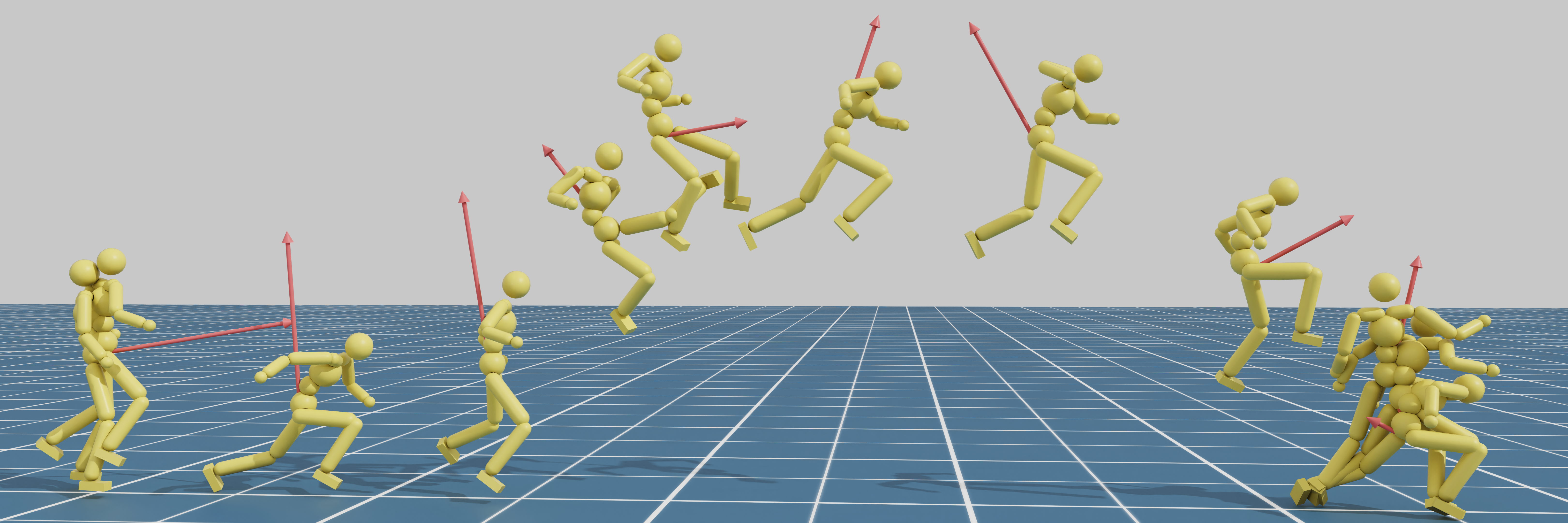}
 \centering
  \caption{Our method enables the reproduction of "physics-defying" anime-style combat skills in a standard physics engine. The sequence illustrates a Dashing Aerial Combat maneuver, featuring instantaneous ground acceleration, a rising kick, and multi-directional mid-air dashes. The visualized vectors (red) represent the learned Assistive Impulse, which injects precise momentum at key kinematic transitions to satisfy the high-dynamic requirements that exceed standard actuation limits.}
\label{fig:teaser}
}

\maketitle
%-------------------------------------------------------------------------
\begin{abstract}
  Physics-based character animation has become a fundamental approach for synthesizing realistic, physically plausible motions. While current data-driven deep reinforcement learning (DRL) methods can synthesize complex skills, they struggle to reproduce exaggerated, stylized motions, such as instantaneous dashes or mid-air trajectory changes, which are required in animation but violate standard physical laws. The primary limitation stems from modeling the character as an underactuated floating-base system, in which internal joint torques and momentum conservation strictly govern motion. Direct attempts to enforce such motions via external wrenches often lead to training instability, as velocity discontinuities produce sparse, high-magnitude force spikes that prevent policy convergence. We propose Assistive Impulse Neural Control, a framework that reformulates external assistance in impulse space rather than force space to ensure numerical stability. We decompose the assistive signal into an analytic high-frequency component derived from Inverse Dynamics and a learned low-frequency residual correction, governed by a hybrid neural policy. We demonstrate that our method enables robust tracking of highly agile, dynamically infeasible maneuvers that were previously intractable for physics-based methods.
%-------------------------------------------------------------------------
%  ACM CCS 1998
%  (see https://www.acm.org/publications/computing-classification-system/1998)
% \begin{classification} % according to https://www.acm.org/publications/computing-classification-system/1998
% \CCScat{Computer Graphics}{I.3.3}{Picture/Image Generation}{Line and curve generation}
% \end{classification}
%-------------------------------------------------------------------------
%  ACM CCS 2012
   % (see https://www.acm.org/publications/class-2012)
%The tool at \url{http://dl.acm.org/ccs.cfm} can be used to generate
% CCS codes.
%Example:


\end{abstract}  
%-------------------------------------------------------------------------
\section{Introduction}\label{sec:introduction}
%New short version is here, old long commented below
% Data-driven reinforcement learning (DRL) has emerged as a way of synthesizing high-fidelity, physically plausible motions by imitating motion references. However, strict adherence to continuous Newtonian dynamics creates an inherent conflict with stylized character animation. Artistic expression often demands highly exaggerated maneuvers - instantaneous acceleration (flash steps), mid-air trajectory changes (double jump), or anti-gravity motions (floating or flying) that explicitly violate the conservation laws, effectively demanding non-physical external propulsion to execute.

The rapid progress in physics-based character animation has demonstrated its effectiveness in creating responsive, physics-plausible motions. Transforming the animation by enabling the agent to interact robustly with the environment offers a level of physical responsiveness that purely kinematic approaches lack. In particular, data-driven reinforcement learning (DRL) has emerged as a way of synthesizing high-fidelity, physically plausible motions by imitating motion references. However, strict adherence to continuous Newtonian dynamics creates an inherent conflict with stylized character animation. Artistic expression often demands highly exaggerated maneuvers - instantaneous acceleration (flash steps), mid-air trajectory changes (double jump), or anti-gravity motions (floating or flying) that explicitly violate the conservation laws, effectively demanding non-physical external propulsion to execute.

Reproducing such behaviors is structurally impossible for existing physics-based controllers. Unlike kinematic methods that can arbitrarily manipulate the global root trajectory, a physically simulated character cannot directly actuate its global position. Instead, it is strictly bonded by contact dynamics, relying on the ground reaction forces and friction to generate the momentum required for locomotion. Consequently, existing learning frameworks are structurally incapable of simulating these physics-infeasible maneuvers as they lack the mechanism to generate the necessary linear and angular momentum without environmental contact.

% Reproducing such behaviors is structurally impossible for existing physics-based controllers, such as AMP and ADD \bb{missing} \cite{todo}. These frameworks model the character as an underactuated floating-base system, which lacks the ability to generate external propulsion. Unlike kinematic methods that can arbitrarily manipulate the global root trajectory, a physically simulated character cannot directly actuate its global position. Instead, it is strictly bonded by contact dynamics, relying on the ground reaction forces and friction to generate the momentum required for locomotion. Consequently, existing learning frameworks are structurally incapable of simulating these physics-infeasible maneuvers as they lack the mechanism to generate the necessary linear and angular momentum without environmental contact.

A seemingly straightforward solution is to introduce explicit assistive wrenches (linear forces and angular torques) applied directly to the character's root, effectively ``actuating'' the base. However, formulating these interventions directly in force space creates an optimization barrier for neural policy learning. Mathematically, the force magnitude required to drive a kinematic discontinuity scales inversely with the time-step ($F \propto 1/\Delta t)$. In the context of exaggerated animation, this manifests as temporally sparse, high-frequency force spikes. These signal characteristics pose a twofold challenge for deep reinforcement learning: first, standard Gaussian exploration strategies struggle to discover statistical outliers in the distribution tails; second, even if sampled, the inherent spectral bias of neural networks~\cite{rahaman2019spectralbiasneuralnetworks} impedes the regression of such high-frequency transients. Consequently, the policy inevitably smoothes the motion's sharpness, failing to converge to the extreme variance in the target signal and thereby losing the instantaneous dynamics of the artistic reference.

We propose Neural Assistive Impulses (NAI), a novel method that shifts from force-based tracking to Momentum-Space Control. By regulating the time-integral of force (impulse) rather than instantaneous force, we transform unbounded force spikes into finite, learnable state transitions. We achieve this through a Hybrid Dynamics Architecture that integrates analytical models with neural residual control. Instead of forcing a neural policy to learn the entire control manifold from scratch, we decompose the actuation into two components: a nominal physical baseline derived analytically via the Recursive Newton-Euler Algorithm (RNEA)~\cite{luh1980online}, and a learnable neural residual impulse. These two streams are dynamically fused to compute the final assistive wrench applied directly to the character's root. This fusion is governed by a Confidence-Aware Dynamics Gate ($\beta$), which functions as a reliability metric: it leverages the RNEA baseline as a directional guidance signal to ensure physically plausible exploration, while adaptively recruiting the Neural Residual to compensate for the data-to-sim discrepancies (e.g. mismatches between RNEA optimization and real-time simulation), thereby executing the physics corrections that the analytical baseline fails to capture. As demonstrated in Fig.~\ref{fig:teaser}, NAI enables complex motion skills such as a rapid rising kick combined with a mid-air dash.

Our approach (NAI) bridges the fundamental gap between kinematic imagination and dynamic reality. By reconciling the intractable discontinuities of exaggerated animation with the stability requirements of physical simulation, we enable the robust reproduction of physics-defying motions within a unified control policy. 
Our technical contributions are (1) \textit{Momentum-Space Control Framework:} We propose a novel control method that resolves the numerical singularities of force-based tracking by formulating exaggerated maneuvers as finite momentum transfers. (2) \textit{Confidence-Aware Hybrid Architecture:} We introduce a dual-stream system governed by a dynamics gate ($\beta$) that synergizes analytical guidance with neural residual learning to robustly bridge data-to-sim discrepancies. (3) \textit{Directional Consistency Constraint:} We propose the Shadow Compass Loss to decouple action direction from magnitude, exploiting analytical baselines for directional regularization to significantly accelerate policy convergence.

\section{Related Work}\label{secl:related-work}
Physics-based character animation has long been a fundamental goal in computer graphics. Physics-based methods focus on designing controllers that actuate the character's motors to interact with a simulated physical world. Early physics-based controllers relied on trajectory optimization and simplified models \cite{Raibert1991-xr,Coros2010-zh,Ye2010-rv,Safonova2004-qc}. These approaches utilize optimization techniques to generate real-time animation while interacting with the environment. With the surge of deep reinforcement learning, more deep learing based methods have become powerful tools for designing such controllers.

\textbf{Deep Reinforcement Learning (DRL)} uses deep neural networks to approximate policy or value functions. This enables the control of complex agents by bypassing the need for explicit analytical modeling of system dynamics within the control loop. Early works applied neural networks for locomotion tasks \cite{Grzeszczuk1995-ew,Grzeszczuk1998-ge}. DRL methods evolved to handle increasingly high-dimensional continuous control problems. Value-based methods demonstrated remarkable capabilities in discrete environments \cite{vanhasselt2015deepreinforcementlearningdouble}, and policy gradient methods proved more effective for the continuous nature of physical locomotion \cite{schulman2017trustregionpolicyoptimization,schulman2017proximalpolicyoptimizationalgorithms,Silver2014DeterministicPG}. In particular, Proximal Policy Optimization (PPO) \cite{schulman2017proximalpolicyoptimizationalgorithms} balances the sample efficiency, stability, and ease of implementation. Recently, DRL has further demonstrated its robustness in challenging scenarios, including rapid training regimes and successful simulation-to-real-world transfer for legged robots \cite{rudin2022learningwalkminutesusing,Hwangbo2019-jc,10403977}.

\textbf{Imitation Learning (IL)} learns complex motion styles from kinematic reference clips while retaining the physical fidelity required for interaction. Tracking-based imitation explicitly designs the objective function to minimize the error between the simulated character and the target poses \cite{Liu2010-wd,Liu2018-gb,2015-TOG-terrainRL,2016-TOG-deepRL,2018-TOG-deepMimic}. However, these methods suffer from the laborious process of ``reward engineering,'' requiring manual tuning of weights and parameters to balance tracking accuracy with physical robustness. Recent works have introduced adversarial learning frameworks that use a discriminator to automatically learn a motion-style reward signal from a dataset of unstructured motion clips \cite{2021-TOG-AMP,zhang2025ADD}. This allows agents to perform tasks while naturally adhering to a stylistic distribution. Recent research hasintroduced hierarchical architectures that separate low-level motor skills from high-level planning, enabling more flexible and interactive user control \cite{2022-TOG-ASE,tessler2023calm,Dou2023-qs,Pan2025-vq}.

\textbf{Exaggerated Animation and Assistive Forces} address a critical demand in film and game production, where characters are often required to perform highly stylized movements that defy the laws of physics. Kinematic approaches for modeling these animations are straightforward, as they allow for the direct manipulation of character poses without dynamic constraints. Consequently, many researchers have focused on generating exaggerated animation efficiently while attempting to preserve a degree of physical plausibility \cite{Wang2006-mq,10.1145/3641519.3657457,xie2025physanimatorphysicsguidedgenerativecartoon}.

Our work aims to bridge this gap by integrating exaggerated animation into a fully simulated physics environment through the use of external assistants. In the context of physics-based character animation, external forces are typically employed as adversarial perturbations to strengthen the robustness of locomotion policies under Newtonian dynamics \cite{Yuan2020-mt,Yu2018-ph,Margolis2022-kc}. In specific domains such as aerial locomotion, prior works have modeled aerodynamic forces to enable the flight of dragon-like characters \cite{Ju2013-zl, Won2018-ms, Won2019-ec}, adapting to various sizes and user interactions. While recent research has utilized external forces to guide the motion trajectory of characters \cite{Kim2025-rq}, these methods often lack the ability to fuse complex, diverse motion skills. distinct from these approaches.
\section{Overview}\label{sec:overview}
\begin{figure*}[t]
\centering
\includegraphics[width=0.99\linewidth]{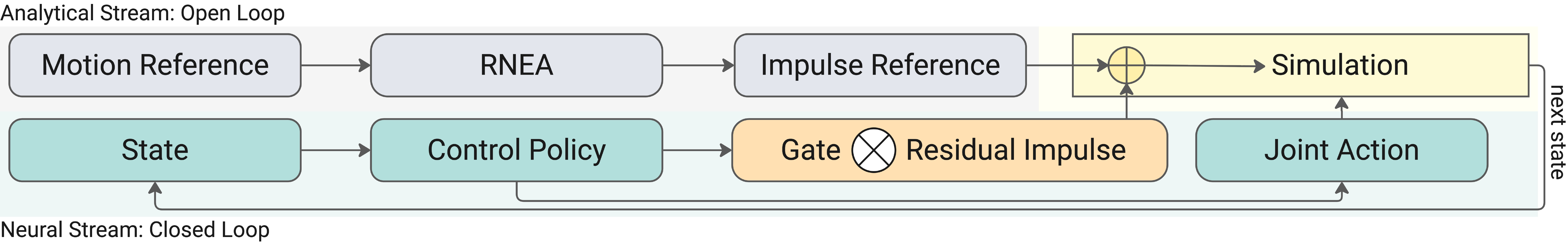}
\caption{\textbf{Overview of the Hybrid Dynamics Architecture:} We decouple the control problem into two parallel streams. The Analytical Stream (Top) serves as an open-loop feed-forward guide, employing an RNEA solver to derive a nominal \textit{Impulse Reference} from the target motion. The Neural Stream (Bottom) operates as a closed-loop feedback controller; the \textit{Control Policy} observes the current simulation \textit{State} and the target reference to predict a \textit{Residual Impulse}, modulated by a learned \textit{Gate}. These components are dynamically fused to drive the \textit{Physics Simulation}, enabling the character to robustly track exaggerated maneuvers that are intractable for purely analytical or purely learning-based methods.}
    \label{fig:overview}

\end{figure*}
Our framework (see Fig.~\ref{fig:overview}) enables a physically simulated character to robustly track highly dynamic, stylized reference motions that may contain kinematic discontinuities or physically infeasible transitions. The system operates via a Hybrid Dynamics Architecture that decouples the control problem into two parallel streams: an analytical physical baseline and a learnable neural residual.

\textit{Input and Preprocessing.}
The system takes a raw motion sequence (e.g., from MoCap or handcrafted kinematic animation) as input. To ensure kinematic compatibility, we first map the source motion to the simulation character's skeleton using optimization-based retargeting~\cite{Zakka_Mink_Python_inverse_2025}, yielding the reference state trajectory $
\mathbf{q}_{ref} = \{q, \dot{q}, \ddot{q}\}.$

\textit{The Analytical Stream.} Given a reference motion, we first treat the character as a fully actuated system and employ the Recursive Newton-Euler Algorithm (RNEA)~\cite{luh1980online} to compute the inverse dynamics. This provides a nominal control signal, specifically, the root assistive wrench required to track the motion under ideal rigid-body assumptions. This stream serves as a guidance signal, ensuring the character follows the general laws of physics and reducing the exploration space for the learning agent.

\textit{The Neural Stream.} While RNEA handles continuous dynamics, it fails when the reference motion violates physical consistency (e.g., sudden velocity jumps or infinite force requirements). To bridge this gap, we introduce a Neural Residual Policy. Instead of predicting forces, this network operates in Momentum Space, predicting a finite-impulse residual to compensate for discrepancies between the RNEA baseline and simulation reality. This allows the system to handle ``impossible'' maneuvers by smoothing out force singularities over discrete time steps.

\textit{Gated Fusion and Execution.} The interplay between these two streams is governed by a Confidence-Aware Dynamics Gate ($\beta$), which dynamically adjusts the reliance on the neural residual based on the tracking error and physical plausibility. The final output is a composite Assistive Wrench (force and torque) applied to the character's root, combined with PD-controlled joint torques. This hybrid formulation allows the character to leverage the stability of analytical models for standard movements while exploiting the plasticity of neural networks for stylized transients.

%In the following sections, we first analyze the theoretical limitations of standard force-based tracking that necessitate this design ~\ref{sec:background}, and then detail the mathematical formulation of our momentum-space control ~\ref{sec:assist-target}. We then detail the neural network architecture at Sec. ~\ref{sec:network} and the geometric training strategy that enables robust policy learning at Sec. ~\ref{sec:optimization}
\section{Assistive Impulse Targets}\label{sec:assist-target}

To robustly reproduce exaggerated maneuvers without inducing simulation instability, we first augment the character's physical model by introducing an external assistive wrench.
However, directly learning this assistive intervention via reinforcement learning presents two fundamental challenges: 1)~High Magnitude Spikes \& Optimization Bias and 2)~Exploration Sparsity.

\textit{High Magnitude Spikes \& Optimization Bias.} 
\begin{figure}[hbt]
    \centering
    \includegraphics[width=0.99\linewidth]{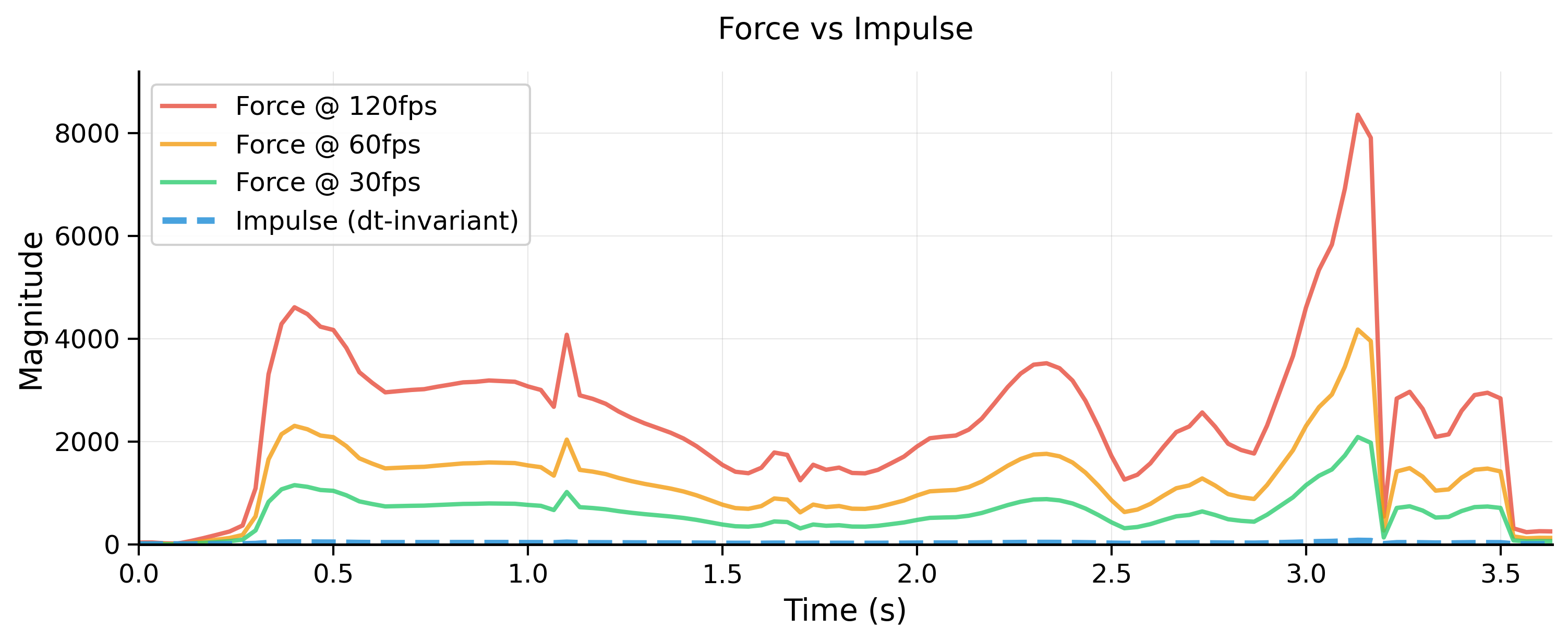}
    \caption{Comparison between Force-space and Momentum-space control signals. We show the magnitude profiles of the assistive intervention for a simulated character performing the motion described in the teaser. While the instantaneous Assistive Force (solid lines) suffers from extreme magnitude spikes and dependency on simulation time-steps ($F \propto 1/\Delta t$), the integrated Assistive Impulse (blue dashed) provides a smooth, bounded, and frequency-invariant learning target for the neural policy.}
    \label{fig:spike}
\end{figure}
The required forces often manifest as sharp, high-frequency spikes. Crucially, the magnitude of these force peaks scales inversely with the simulation time-step ($F \propto 1/\Delta t$), resulting in unbounded signals that diverge as temporal resolution increases (e.g., exceeding 4kN at 60Hz, and 8kN at 120Hz)(See Fig. ~\ref{fig:spike}). Neural networks struggle to approximate these signals due to spectral bias, which prioritizes the learning of low-frequency functions~\cite{rahaman2019spectralbiasneuralnetworks}. 
Furthermore, the high sensitivity of the control problem exacerbates this; even small biases in force prediction can lead to dramatic errors in motion tracking.
% Removed parentheses and "Note:" for better academic flow
Although logarithmic scaling could theoretically compress these values, it violates the linear superposition principle required for our residual learning formulation (see Sec.~\ref{sec:network} for details). Consequently, we project the target into Momentum Space, transforming intractable force spikes into physically bounded, stable momentum transfers.

\textit{Exploration Sparsity.} Even with a stable action space, valid assistive impulses are typically temporal outliers. They activate only within extremely narrow time windows while remaining silent for the rest of the motion cycle. Standard random exploration strategies struggle to sample these values, as the optimal impulses lie in the distribution's extreme tails.

We do not learn the assistive impulse from scratch. Instead, we compute an Analytical Baseline ($\mathbf{I}_{base}$) from inverse dynamics to guide exploration, and train the policy to predict only a Residual Impulse ($\mathbf{I}_{res}$). This allows the network to focus on correcting local discrepancies rather than discovering global dynamics. The final assistive impulse $I_{assistive}$ applied to the character is:
\begin{equation}
    I_{assistive} = I_{baseline} + I_{residual}
\end{equation}

\subsection{Floating Base Dynamics}
We begin by establishing the fundamental floating-base dynamics that govern our character to analytically derive this baseline. The dynamic system of our character with $n$ joints in a floating base system is modeled as:
\begin{equation}
M(q)\dot{v} + C(q,v) = S^T \tau + J_c^T f_c + \mathbf{W_{assist}}
\label{eq:floating-base-dynamics}
\end{equation}
where $q = [q_{base}, q_{joints}]^T$ denotes the generalized coordinates, with $q_{base}$ parameterizes the floating base pose in $SE(3)$ and $q_{joints} \in \mathbb{R}^n$ represents the configuration of $n$ actuated joints, $v \in R^{6+n}$ denotes the generalized velocity vector, $M(q)$ and $C(q,v)$ denote the mass-inertia matrix and the generalized bias force vector and $\tau \in \mathbb{R}^n$ denotes the vector of internal actuation torques. The selection matrix $S^T = [\mathbf{0}_{n \times 6}, \mathbf{I}_{n \times n}]^T$ selects the actuated degrees of freedom, explicitly enforcing the underactuation of the base. $J_c^T f_c$ accounts for contact forces. 
Finally, $\mathbf{W}_{assist} \in \mathbb{R}^{n+6}$ represents the \textbf{assist wrench} applied to the base. The external wrench serves to bridge the ``dynamics gap'' when ground contact forces and internal actuation are insufficient to track the target motion.

\subsection{Inverse Dynamics Analysis}\label{subsec:inv_dynamics}
To compute the necessary assistance, we employ a two-stage inverse dynamics pipeline. In this subsection, we focus on the first stage: determining the net dynamic demand. The subsequent decomposition of this demand (Stage 2) follows in Sec ~\ref{ssec:decomposition}.

Since the reference motion consists of discrete poses $\hat{q}_{t}$, we first extract the required velocities $\hat{v}$ and accelerations $\hat{\dot v}$ from the reference.
With the target kinematic $(\hat q, \hat v, \hat{\dot v})$ from the motion reference, we compute the total generalized wrench $\tau_{req}$ sustain the motion.
Using the Recursive Newton-Euler Algorithm (RNEA)~\cite{luh1980online}, we obtain: \begin{equation} 
\tau_{req} = \text{RNEA}(\hat{q}, \hat{v}, \hat{\dot{v}}) = M(\hat{q})\hat{\dot{v}} + C(\hat{q},\hat{v}). 
\end{equation}

Note that $\tau_{req} \in \mathbb{R}^{n+6}$ represents the total physical demand (the ``net force'') required to satisfy the equations of motion. It aggregates all inertial forces, gravity, and Coriolis effects required to satisfy the equations of motion.
At this stage, this demand is treated as a unified vector; the specific allocation between ground contacts and assistive wrench is resolved in the next step.

\subsection{Assistive Wrench Decomposition}\label{ssec:decomposition}
Next, we decompose this total demand to isolate the minimal assistive intervention.
Since the character is underactuated, the floating base cannot generate internal torque. The total demand~$tau_{req}$ should be satisfied with the mixing of ground contact force and assistive wrench. We express this decomposition as:
\begin{equation}
    \tau_{req} = S^T\tau_{motor} + J_c^T f_c+ \mathbf{W}_{assist}.
\end{equation}

Focusing specifically on the unactuated base coordinates (where internal torque is zero, i.e., $S^T\tau{motor} = \mathbf{0}$), the equation simplifies :
\begin{equation}
\boldsymbol{\tau}^{base}_{req} = \mathbf{J}_{c,base}^T f_c + \mathbf{W}_{assist},
\end{equation}
where $\boldsymbol{\tau}^{base}_{req} \in \mathbb{R}^6$ denotes the spatial force and torque requirement at the root derived from $\tau_{req}$.

To resolve the redundancy in the force distribution problem (i.e., deciding how much comes from the ground and the external assistance), we formulate it as a Quadratic Programming (QP) problem at each time step. The optimization minimizes the assistive intervention with the necessary maximum ground reaction force:
\begin{equation}
\begin{aligned}
\min_{\mathbf{f}_c, \mathbf{W}_{\text{assist}}} \quad & \|\mathbf{W}_{\text{assist}}\|_{\mathbf{Q}}^2 + \lambda \|\mathbf{f}_c\|^2 \\
\text{s.t.} \quad & \mathbf{J}_{c, \text{base}}^\top \mathbf{f}_c + \mathbf{W}_{\text{assist}} = \boldsymbol{\tau}^{base}_{req} \\
& \mathbf{f}_{c, i} \in \mathcal{K}_{\mu}, \quad \forall i \in \mathcal{C},
\end{aligned}
\label{eq:qp_target}
\end{equation}
where, $\mathbf{Q} \in \mathbb{R}^{6 \times 6}$ is a positive semi-definite weight matrix, and $\lambda$ is a regularization coefficient. The set $\mathcal{C}$ contains the indices of active contacts, and $\mathcal{K}_{\mu}$ denotes the linearized Coulomb friction cone with friction coefficient $\mu$.

Critically, we assign a higher penalty weight to the vertical component of $\mathbf{W}_{assist}$ within $\mathbf{Q}$. This encourages the solver to maximize the use of ground reaction forces for support, activating the assistive wrench only when physical contacts are insufficient to track the motion (e.g., during physically infeasible flight phases). The optimized wrench $\mathbf{W}_{assist}$ derived from Eq.~\ref{eq:qp_target} represents an instantaneous force. Using this sparse, high-frequency signal directly as a learning target is numerically unstable.

Therefore, we explicitly project this wrench into momentum space by integrating it over the simulation timestep $\Delta t$. We define the final Analytical Impulse Baseline $\mathbf{I}_{base}$ as:
\begin{equation}
\mathbf{I}_{base} = \int{t}^{t+\Delta t} \mathbf{W}{assist}(\tau) d\tau \approx \mathbf{W}{assist} \cdot \Delta t
\end{equation}
This transformation serves two critical purposes: it normalizes the intractable force spikes into physically bounded momentum transfers, and it aligns the baseline dimensionality with our residual learning framework. The computed $\mathbf{I}_{base}$ is then passed to the control policy, which learns to predict the residual correction $\mathbf{I}_{res}$ (Sect.~\ref{sec:network}).
\section{Neural Residual Impulse Control}\label{sec:network}
The Analytical Impulse Baseline ($\mathbf{I}_{base}$) derived in Sec.~\ref{sec:assist-target} provides a physically grounded estimate of the necessary intervention. However, it is fundamentally a \textbf{time-indexed open-loop signal}, computed based on the ideal reference trajectory.
Directly replaying this baseline in a dynamic simulation is prone to failure due to the open-loop brittleness problem. 
Specifically, the forward simulation involves discrete numerical integration and iterative constraint-solving (e.g., friction constraints, penetration recovery), which inevitably introduce deviations from the ideal inverse dynamics model. These deviations manifest in two critical ways:
\begin{itemize}
    \item \textit{Temporal Desynchronization}: The character in simulation often lags behind or leads the reference motion due to inertia or contact delays. If the character is delayed by even a few frames, the pre-calculated $\mathbf{I}_{base}$, which might encode a massive take-off impulse, will trigger at the wrong kinematic phase (e.g., while the character is still crouching), destabilizing rather than assisting the motion.
    \item \textit{State-Action Divergence}: Small errors in root orientation or velocity accumulate rapidly over time. Since $\mathbf{I}_{base}$ is ``blind'' to the character's current state, it cannot adapt to these drifts. Applying a fixed force vector to a slightly tilted character generates unintended torque, exacerbating the error instead of correcting.
\end{itemize}
To resolve these numerical and physical limitations, the system must transition from open-loop execution to closed-loop feedback control. We introduce a neural policy $\pi_\theta$ that outputs internal joint targets alongside a residual impulse ($\mathbf{I}_{res}$) to dynamically adapt the analytical baseline to the runtime simulation state. The final root control law is formulated as:
\begin{equation}
    \mathbf{I}_{total} = \mathbf{I}_{base}(\phi,s_t) + \mathbf{I}_{res}(s_t),
    \label{eq:residual_control}
\end{equation}
\begin{equation}
    \mathbf{I} = \mathbf{F} \Delta t
    \label{eq:impulse_integrate}
\end{equation}
where the neural term $\mathbf{I}_{res}(s_t)$ functions as a stabilizing feedback controller. It mathematically compensates for the dynamics gap caused by discrete solver integration errors and realigns the applied impulse with the character's instantaneous physical state ($s_t$). This formulation transforms the unstable open-loop approximation into a robust closed-loop control system.

\subsection{Network Architecture}\label{ssec:architecture}
\begin{figure}[hbt]
    \centering
    \includegraphics[width=0.99\linewidth]{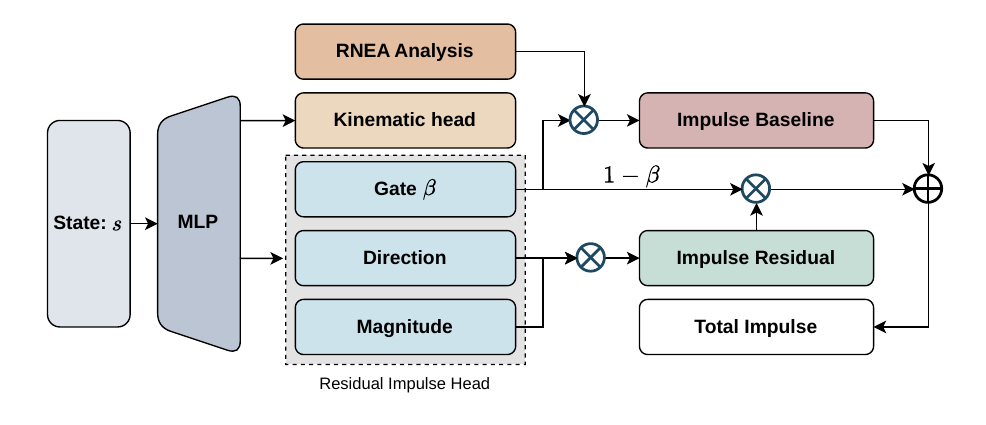}
    \caption{The detailed architecture of the proposed Neural Assistive Impulse (NAI) policy network.}
    \label{fig:network}
\end{figure}

We parameterize the control policy $\pi_\theta(\mathbf{a}_t | \mathbf{s}_t)$ using a dual-head neural network, the structural details of which are illustrated in Figure \ref{fig:network}. To ensure numerical stability under high-dynamic-range impulses, we apply logarithmic feature scaling to the input state and employ a direction-magnitude decomposition for the action space.

\textit{State Representation.} The policy observation $\mathbf{s}_t$ must encapsulate both the kinematic state of the character and the interaction history. 
We adopt the standard proprioceptive observation set utilized in \cite{2018-TOG-deepMimic}, which has been widely followed by subsequent frameworks such as \cite{2021-TOG-AMP, 2022-TOG-ASE, zhang2025ADD}. The observations comprise the root's height, orientation, and linear/angular velocities, along with the local positions and rotations of each body part relative to the root and velocities of all articulated joints. To explicitly inform the policy about the external forcing, we augment the observation with the history of applied impulses.

A critical challenge in encoding impulse information is the vast dynamic range of assistive impulses, which can range from $0\text{N}$ to over $4\text{kN}$. Direct linear input of such values causes feature dominance. To mitigate this, we compress the history of assistive interventions using a signed-logarithmic transformation. We define the state tuple as $\mathbf{s}_t = \{\mathbf{s}_{prop}, \mathbf{s}_{task}, \mathcal{T}(\mathbf{H}_{assist})\}$, where the transformation $\mathcal{T}$ is defined as:
\begin{equation}
    \mathcal{T}(\mathbf{x}) = \text{sgn}(\mathbf{x}) \odot \log(1 + |\mathbf{x}|).
\end{equation}
This scaling preserves the signal's zero crossings and directionality while mapping its magnitude to a tractable range for the network.

\textit{Action Decomposition}
The output layer branches into two specialized heads: a \textbf{Kinematic Head} for pose tracking and a \textbf{Residual Impulse Head} for assistive correction. Recognizing that translation and rotation often require distinct dynamic adjustments, we explicitly decouple the residual output into a \textit{linear impulse} $\mathbf{I}_{lin}$ and an \textit{angular impulse} $\mathbf{I}_{ang}$. To bridge the gap between normalized network outputs and physical world magnitudes, we introduce two hyperparameters, $\sigma_{lin}$ and $\sigma_{ang}$, representing the maximum allowable impulse capacity. The final residual actions are formulated as:
\begin{align}
    \mathbf{I}_{lin} &= \sigma_{lin} \cdot m_{lin} \cdot \mathbf{u}_{lin} \nonumber\\
    \mathbf{I}_{ang} &= \sigma_{ang} \cdot m_{ang} \cdot \mathbf{u}_{ang} \nonumber\\
    \mathbf{I}_{res} &= [\mathbf{I}_{lin}, \mathbf{I}_{ang}],
\end{align}
where the components are defined as follows:
\begin{itemize}
    \item \textit{Direction ($\mathbf{u}_{lin}, \mathbf{u}_{ang} \in S^2$):} Independent unit vectors representing the spatial orientation of the impulses. We constrain the raw network outputs to $[-1, 1]$ via $\tanh$ activation before normalizing them to the unit sphere, ensuring numerical stability.
    \item \textit{Magnitude ($m_{lin}, m_{ang} \in [0, 1]$):} Scalar intensity factors that determine the strength of the assistance. We map the raw outputs to the range $[0, 1]$ to represent the percentage of the maximum capacity.
    \item \textit{Global Scales ($\sigma_{lin}, \sigma_{ang}$):} Constant scaling factors that define the physical limits of the assistance. Generally, we set $\sigma_{lin} = 25\,\text{N} \cdot \text{s}$ and $\sigma_{ang} = 8\,\text{Nm} \cdot\text{s}$ to provide sufficient residual momentum for highly agile maneuvers.
\end{itemize}
Finally, the learned residual impulse $\mathbf{I}_{res}$ is directly added to the analytical baseline derived from RNEA, yielding the total assistive action applied to the character's root.

\subsection{Composite Control Synthesis}
The final control signal applied to the character is synthesized by fusing the modulated analytical baseline with the learned assistive corrections. This formulation serves as a dual-gated arbitration, empowering the policy to dynamically rebalance its trust between the physics-based prior and its self-generated interventions. 

To account for the distinct dynamic ranges of translational and rotational motion, we decouple the gating mechanism into separate linear and angular components. Let $\beta_{lin}, \beta_{ang} \in [0, 1]$ denote the gating scalars output by the neural policy. The total applied impulse $\mathbf{I}_{total} \in \mathbb{R}^6$ is computed via the following complementary block-vector formulation:
\begin{equation}
    \mathbf{I}_{total} = 
    \begin{bmatrix}
    \beta_{lin} \mathbf{I}_{base}^{lin} \\
    \beta_{ang} \mathbf{I}_{base}^{ang}
    \end{bmatrix}
    +
    \begin{bmatrix}
    (1 - \beta_{lin}) \mathbf{I}_{res}^{lin} \\
    (1 - \beta_{ang}) \mathbf{I}_{res}^{ang}
    \end{bmatrix},
    \label{eq:composite_impulse}
\end{equation}
where $\mathbf{I}^{lin} \in \mathbb{R}^3$ and $\mathbf{I}^{ang} \in \mathbb{R}^3$ denote the translational and rotational sub-vectors of the respective $6$D impulses. $\mathbf{I}_{res}$ represents the residual impulse generated by the neural policy. This formulation strictly bounds the interpolation between the analytical baseline and the neural residual for both spatial domains independently.

Since the physics engine integrates forces and torques over discrete time steps, the composite impulses must be mapped to physical wrenches. For a control step $\Delta t$, the effective wrench $\mathbf{W}$ (comprising the external force $\mathbf{F}_{ext}$ and torque $\mathbf{\tau}_{ext}$) applied to the character's root is defined as:
\begin{equation}
\mathbf{W} = \frac{\mathbf{I}_{total}}{\Delta t}.
\end{equation}
This transformation ensures that the policy injects precise momentum increments into the simulator, effectively bridging the gap between kinematic reference trajectories and the underlying physical constraints.

By structurally separating the baseline trust ($g_{base}$) from the residual intervention ($g_{res}$ hidden within $\mathbf{I}_{res}$), the framework encourages emergent sparsity. During physically plausible locomotion, the policy learns to rely on the modulated baseline while keeping the residual channel dormant. Assistive impulses are activated only when the required dynamics exceed the analytical model's capacity, such as during the instantaneous momentum shifts required for high-agility combat maneuvers.

\subsection{Physics-Regularized Optimization}
The policy training is adopted from \cite{zhang2025ADD}, which uses an adversarial learning process to evaluate a reward that balances multiple learning objectives.
Optimizing the composite control law in Eq.~\ref{eq:composite_impulse} presents a coordination challenge. A standard reinforcement learning objective (e.g., PPO) often struggles to sample from the tail of the distribution, leading to slow convergence or degenerate behaviors in which the policy abuses the residual impulse to bypass physical constraints. 

To guide the learning process, we introduce two physics-informed auxiliary objectives: a Shadow Compass Loss to accelerate directional exploration, and an Intervention Sparsity Loss to enforce the minimal assistance principle. The total optimization objective is formulated as:
\begin{equation}
    \mathcal{L}_{total} = \mathcal{L}_{PPO} + w_{c} \mathcal{L}_{compass} + w_{s} \mathcal{L}_{sparsity}.
\end{equation}

Learning the optimal direction $\mathbf{u}$ for the residual impulse from scratch is inefficient, as the reward signal provides only sparse feedback for 3D orientation. We leverage the analytical baseline from Sec.~\ref{sec:assist-target} as a dense supervisory signal, encouraging the learned residual to align with the RNEA-derived force vector $\mathbf{F}_{ref}$. However, a na\"ive alignment is unstable: when the reference force is negligible (e.g., during static phases), its direction is dominated by numerical noise, leading to gradient divergence.

To address this, we employ a Masked Cosine Alignment in which supervision is strictly conditioned on signal intensity. We term this the ``Shadow Compass'' to reflect this adaptive behavior as it provides a directional reference that ``shadows'' the physical intent, yet is effectively ``shadowed'' (masked) itself when the reference signal fades into ambiguity. We employ a Masked Cosine Alignment as follows:
\begin{equation}
    \mathcal{L}_{compass} = \mathbb{E} \left[ 1 - \cos(\mathbf{u}, \mathbf{d}_{target}) \right],
\end{equation}
where the target direction $\mathbf{d}_{target}$ switches dynamically based on the reference intensity:
\begin{equation}
    \mathbf{d}_{target} =
        \begin{cases}
            \mathbf{F}_{ref} / ||\mathbf{F}_{ref}|| & \text{if } ||\mathbf{F}_{ref}|| > \epsilon\\
            \mathbf{u}_{up} & \text{otherwise}
        \end{cases}
\end{equation}
where $\epsilon$ is a noise threshold and $\mathbf{u}_{up} = [0, 0, 1]^\top$. This mechanism aligns the residual with the physics-based intent during dynamic maneuvers, while defaulting to a gravity-opposing vertical bias during quasi-static states, ensuring consistent gradient flow.

To prevent the policy from generating excessive "ghost forces" that violate physics plausibility, we explicitly penalize the activation of the residual head.
We formulate a regularization term that encourages the policy to default to the analytical baseline:
\begin{equation}
    \mathcal{L}_{sparsity} = \lambda_m |m|^2 + \lambda_g |g_{base}|^2
\end{equation}
The first term minimizes the residual magnitude $m$, forcing the network to keep corrections close to zero unless necessary. The second term anchors the baseline gate $g_{base}$ towards $0.0$, encouraging the policy to explore its own impulse. Together, these constraints ensure that the learned residuals emerge only as necessary corrections to bridge the dynamics gap, rather than replacing the physical baseline.

%Collectively, this physics-regularized framework harmonizes the robustness of analytical priors with the adaptability of deep learning. We detail the concrete implementation parameters and training protocol in Sec.~\ref{sec:implementation}.
\section{Implementation}\label{sec:implementation}
We simulate a 28-DOF humanoid character in NVIDIA Isaac Gym \cite{makoviychuk2021isaac}, leveraging massively parallel simulation across 4,096 environments. The physics simulation operates at $60\,\text{Hz}$ and the control policy at $30\,\text{Hz}$. Kinematic actions are converted into joint torques via a stable Proportional-Derivative (PD) controller. The learned residual forces and torques are applied as additive corrections directly to the character's root and joints. The policy is optimized using PPO~\cite{schulman2017proximalpolicyoptimizationalgorithms}, with advantages computed via Generalized Advantage Estimation (GAE)~\cite{schulman2018highdimensionalcontinuouscontrolusing}. The value function is updated by regressing target values computed with Temporal Difference (TD) learning. Training takes approximately 70 million sample steps and requires roughly 8 hours on a NVIDIA GeForce RTX 4090 GPU.

\subsection{Motion Reference Data}
\begin{figure}[ht]
    \centering
    \includegraphics[width=0.99\linewidth]{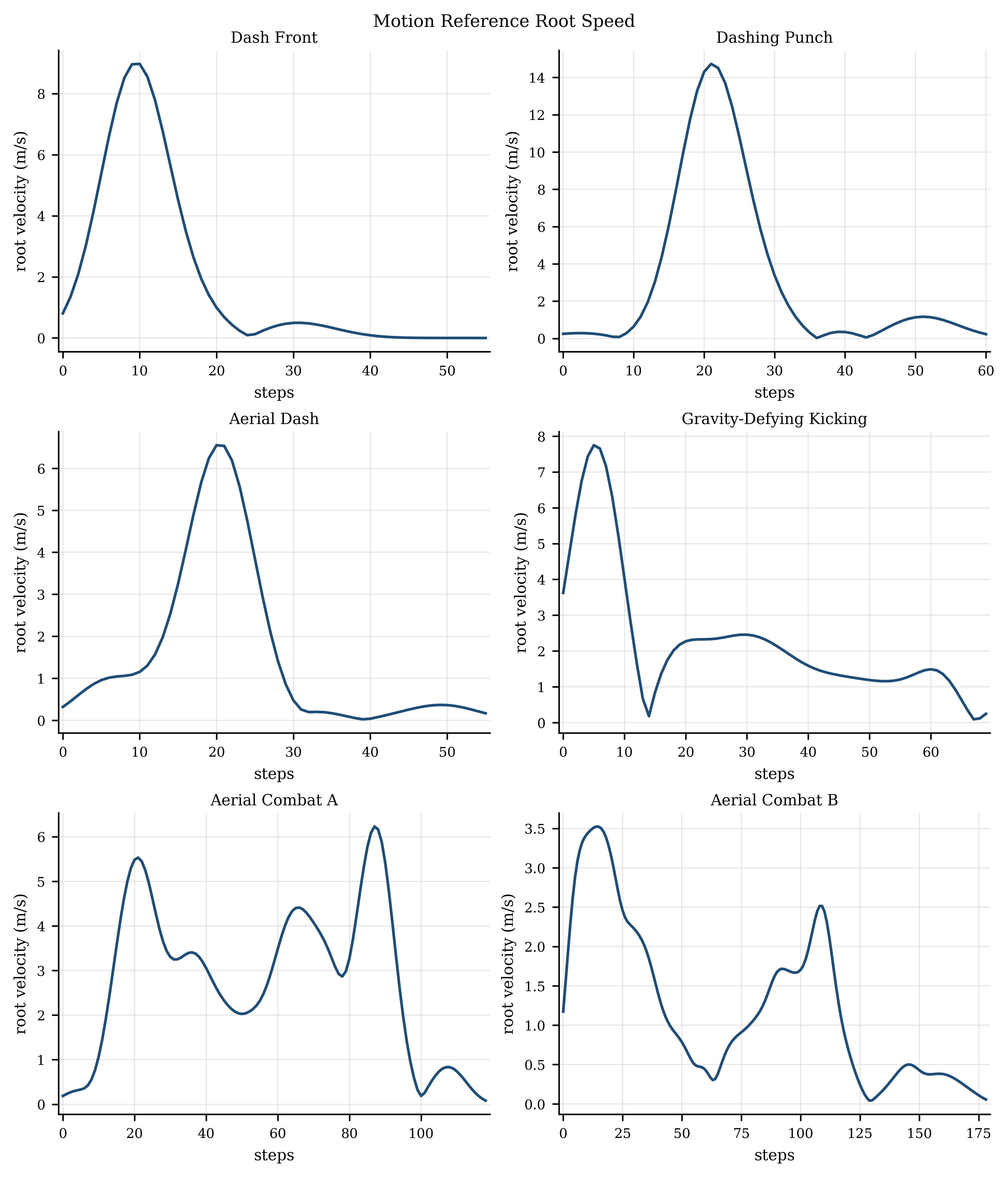}
    \caption{Snapshots of the simulated humanoid characters trained using our NAI framework, showcasing exaggerated and stylized motion capabilities.}
    \label{fig:root_velocity}
\end{figure}

\begin{figure*}[hbt]
    \centering
    \includegraphics[width=0.99\linewidth]{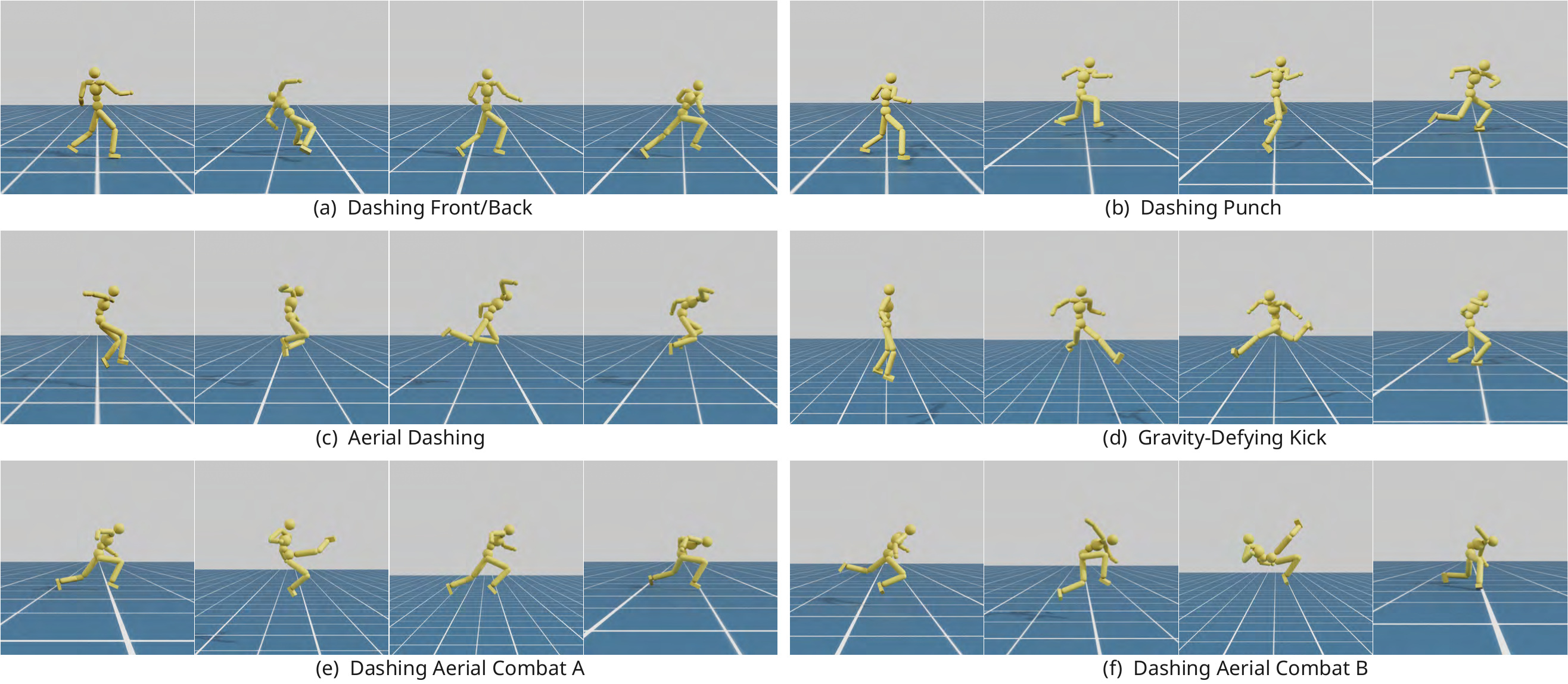}
    \caption{Snapshots of the simulated humanoid characters trained using our NAI framework, showcasing exaggerated and stylized motion capabilities.}
    \label{fig:tasks}
\end{figure*}

To evaluate our framework's capacity to track highly dynamic and stylized behaviors, we leverage the Fight Animations Pack, a commercial library containing over 239 high-fidelity motion-capture and handcrafted clips. A significant challenge with these assets is their short duration (typically $0.2\text{--}1.0\,\text{s}$). To construct meaningful long-horizon control tasks that cover diverse dynamic regimes, we categorize the data into atomic primitives and synthesize them into complex composite motion sequences. (See Fig. ~\ref{fig:tasks})

\textit{Atomic Primitives.} We select a set of foundational high-agility skills that require rapid momentum changes: 
\begin{enumerate} 
    \item \textit{Ground Dashing: }High-speed sliding dashes (forward / backward) with a boost acceleration. 
    \item \textit{Dashing Punch:}  Rapid ``dash-and-stop'' attacks that demand instantaneous acceleration and deceleration. 
    \item \textit{Aerial Dashing:} Mid-air horizontal impulse generation without ground leverage.
\end{enumerate}

\textit{Composite Sequences.} We manually splice these primitives to create physically impossible motion sequences that serve as rigorous stress tests for our residual force generation:
\begin{enumerate}
    \item \textit{Gravity-Defying Kick:} A vertical leap reaching $3.3\,\text{m}$, followed by a controlled slow-motion descent ($< 9.8\,\text{m/s}^2$) while performing combat strikes.
    \item \textit{Dashing Aerial Combat A:} A ground dashing transitioning into a fast kick rising to the air, following up with a mid-air double dash, then landing. (see Fig. ~\ref{fig:teaser})
    \item \textit{Dashing Aerial Combat B:} A ground-to-air dashing transitioning into a mid-air double jump (continuing to rise during a kick), culminating in a high-velocity downward smash.
\end{enumerate}

The composite sequences require the policy to switch between ground contact utilization and pure residual-driven aerial maneuvering, validating the system's stability across varied contact states.

To quantitatively demonstrate the kinematic difficulty of these stylized skills, we visualize the root velocity profiles across the dataset (see Figure \ref{fig:root_velocity}). The trajectory data reveals that these motions frequently exhibit high-magnitude peak velocities and near-instantaneous velocity step changes. This concentration of momentum variation formally validates that tracking these specific sequences constitutes a dynamically ill-posed problem for strictly underactuated systems, thereby motivating the necessity of our residual impulse formulation.

\subsection{Network Architecture}
Both the actor and critic networks are parameterized as MLPs. The \textit{Actor} network maps the input state $\mathbf{s}_t$ through two hidden layers of $[1024, 512]$ units with ReLU activations~\cite{agarap2019deeplearningusingrectified}. The output layer branches into two heads: (1) The Kinematic Head outputs a 28-dimensional vector parameterizing a diagonal Gaussian distribution $\pi_\theta(\mathbf{a}_t|\mathbf{s}_t) = \mathcal{N}(\mu_t(\mathbf{s}_t), \Sigma)$ for joint targets. (2) The Residual Head outputs a 10-dimensional vector, where the direction $\mathbf{u}$ is normalized after a $\tanh$ activation, while the magnitude $m$ and gate $g$ are mapped to $[0, 1]$ via a scaled $\tanh$ function.

The \textit{Critic} network shares the same hidden layer structure $[1024, 512]$ but outputs a single scalar value estimate $V(\mathbf{s}_t)$. Additionally, the Adversarial Differential Discriminators (ADD) utilize a Discriminator network (structure $[1024, 512]$) to distinguish between simulated transitions and the reference dataset.

\section{Experiments}\label{sec:experiments}

\subsection{Baseline Validation}
\begin{table}[htbp]
  \centering
  \caption{Comparison of Tracking Success Rate Across Exaggerated Motion Categories. The standard underactuated baseline (ADD/AMP) exhibits a 0\% success rate across all test sequences, as pure joint torques are mathematically insufficient to execute the required momentum discontinuities. Our proposed NAI framework achieves a 100\% success rate across the identical test set.}
  \label{tab:baseline_success_rate}
  \begin{tabular}{lcc}
    \toprule
    \textbf{Motion Sequence} & \textbf{ADD/AMP (Baseline)} & \textbf{NAI (ours)} \\
    \midrule
    Dashing Attack           & 0\%                              & 100\%                    \\
    Ground Dashing           & 0\%                              & 100\%                    \\
    Aerial Dashing           & 0\%                              & 100\%                    \\
    Multi-Dir Combat         & 0\%                              & 100\%                    \\
    Aerial Launcher          & 0\%                              & 100\%                    \\
    Gravity-Defying          & 0\%                              & 100\%                    \\
    \bottomrule
  \end{tabular}
\end{table}

To quantitatively evaluate the capability of existing state-of-the-art (SOTA) physics-based character control frameworks in handling exaggerated motions, we establish baseline comparisons using Adversarial Motion Prior (AMP) \cite{2021-TOG-AMP} and Adversarial Differential Discriminators (ADD) \cite{zhang2025ADD} architectures. In these baseline models, the control policy is constrained to a strictly underactuated physical formulation, meaning the root link is strictly unactuated, and no external assistive virtual forces or impulses are applied.

\noindent{\textbf{Evaluation Metrics:}} The primary metric for this baseline validation is the \textit{Success Rate}. It is quantitatively defined as the percentage of evaluation episodes wherein the character tracks the reference kinematic trajectory for the entire sequence duration without triggering termination conditions (i.e., falling or exceeding a predefined maximum root position error threshold). The success rate is calculated over 4096 environments.

\noindent{\textbf{Results and Analysis:}} The quantitative results demonstrate a categorical failure of the standard baselines when confronted with time-domain discontinuities. Across all five tested exaggerated motion sequences, the ADD/AMP baselines yielded a Success Rate of exactly 0\%. In contrast, our proposed momentum-space neural control framework achieved a 100\% Success Rate across the identical test set. 

As documented in Table \ref{tab:baseline_success_rate}, this systematic failure occurs consistently across all tested motion categories. The baseline models trigger termination conditions immediately following the onset of non-physical motion segments, such as instantaneous spatial translations or mid-air accelerations. This baseline failure is theoretically anticipated; it does not invalidate the efficacy of ADD/AMP in tracking physically valid regular motions. Rather, it provides empirical evidence that the external assistive impulse is strictly necessary for executing kinematically exaggerated motions that violate momentum conservation.

\subsection{Quantitative Tracking Fidelity Comparison}

\begin{table*}[t]
\centering
\footnotesize
\caption{Quantitative comparison against the baseline. We report Position (Pos) and Velocity (Vel) errors, followed by Impulse analysis. Each task compares our \textbf{Dual-Gated} method against the \textit{Naive Baseline}. (N/A indicates unavailable baseline data). All the motion skills are evaluated with 4096 }
\label{tab:quantitative_results}

\begin{tabular*}{\textwidth}{l l @{\extracolsep{\fill}} cc cc cc cc cc c}
\toprule
\textbf{Task} & \textbf{Method} & \multicolumn{2}{c}{\textbf{Pos Error} [m]} & \multicolumn{2}{c}{\textbf{Vel Error} [rad/s]} & \multicolumn{2}{c}{\textbf{Total Imp. $\downarrow$}} & \multicolumn{2}{c}{\textbf{Ref Imp.}} & \multicolumn{2}{c}{\textbf{Res Imp.}} & \textbf{Jitter $\downarrow$} \\
\cmidrule(lr){3-4} \cmidrule(lr){5-6} \cmidrule(lr){7-8} \cmidrule(lr){9-10} \cmidrule(lr){11-12} \cmidrule(lr){13-13}
& & Mean & Std & Mean & Std & Lin  & Ang & Lin & Ang & Lin & Ang & [Unit] \\
\midrule
% Task 2
Ground Dashing & \textbf{NAI} & 0.022 & 0.032 & 0.42 & 0.064 & \textbf{15.92} & 4.93 & 5.51 & 3.82 & 11.53 & 2.18 & \textbf{8.7} \\
& Naive & 0.007 & 0.002 & 0.161 & 0.060 & 20.64 & \textbf{2.43} & N/A & N/A & N/A & N/A & 14.7 \\

\midrule
% Task 1
Dashing Punch & \textbf{NAI} & 0.020 & 0.039 & 0.48 & 0.112 & \textbf{34.93} & 8.84 & 21.50 & 7.60 & 16.92 & 2.72 & \textbf{8.84} \\
& Naive & 0.031 & 0.031 & 0.277 & 0.116 & 36.54 &  \textbf{2.43} & N/A & N/A & N/A & N/A & 10.91 \\
\midrule

% Task 3
Aerial Dashing & \textbf{NAI} & 0.013 & 0.006 & 0.17 & 0.070 & \textbf{21.46} & \textbf{2.30} & 22.56 & 3.13 & 0.86 & 2.27 & \textbf{3.6} \\
& Naive & 0.005 & 0.002 & 0.175 & 0.069 & 23.24 & 2.51 & N/A & N/A & N/A & N/A & 13 \\
\midrule

% Task 4
Dashing Aerial Combat A & \textbf{NAI} & 0.015 & 0.007 & 0.19 & 0.090 & \textbf{19.65} & 3.85 & 12.53 & 3.53 & 8.56 & 2.80 & \textbf{6.60} \\
& Naive & 0.005 & 0.003 & 0.153 & 0.065 & 22.5 & \textbf{2.89} & N/A & N/A & N/A & N/A & 22.50 \\
\midrule

% Task 5
Dashing Aerial Combat B & \textbf{NAI} & 0.073 & 0.005 & 0.43 & 0.47 & \textbf{14.56} & 3.54 & 10.06 & 2.46 & 5.23 & 1.85 & \textbf{2.4} \\
& Naive & 0.0075 & 0.0045 & 0.166 & 0.1091 & 18.67 & \textbf{3.03} & N/A & N/A & N/A & N/A & 4.2 \\
\midrule

% Task 6
Gravity-Defying Kick & \textbf{NAI} & 0.056 & 0.068 & 0.17 & 0.07 & \textbf{19.84} & \textbf{4.84} & 11.93 & 3.92 & 8.01 & 0.58 & \textbf{4.7} \\
& Naive & 0.084 & 0.096 & 0.433 & 0.330 & 28.2 & 7.63 & N/A & N/A & N/A & N/A & 26.39 \\

\bottomrule
\end{tabular*}
\end{table*}

We evaluate the performance of our framework across 6 skill tasks. The primary objective of this quantitative analysis is to demonstrate that our gated residual architecture achieves \textbf{competitive tracking fidelity} while strictly adhering to the principle of \textbf{minimal physical intervention}, unlike naive approaches that brute-force kinematic alignment via excessive external forces.

\noindent{\textbf{Evaluation Metrics:}} To isolate the impact of our dual-gating and sparsity mechanisms, we compare our method against a \textbf{Native Assistive Impulse} method. 
This implementation uses a network to learn an assistive impulse (6D) based on the ADD architecture but lacks a reference-impulse baseline, gating heads, and sparsity-driven loss functions. In this unregulated setup, the residual forces are ``always-on'' and directly regressed by the policy.

Following ADD~\cite{zhang2025ADD}, we employ three key metrics. We prioritize the balance between tracking error and impulse magnitude, rather than minimizing kinematic error at all costs:
\begin{itemize}
    \item \textit{Body Pose Error ($E_{pose}$):} The root-relative mean squared error (RMSE) of all joint positions, representing kinematic accuracy.
    \item \textit{DoF Velocity Error ($E_{vel}$):} The RMSE of all articulated joint velocities. Lower values indicate smoother, less jittery motion.
    \item \textit{Average Impulse ($\bar{I}$):} The mean magnitude of the applied linear and angular residual impulses ($N \cdot s$) quantifies the ``physical cost'' or violation of standard dynamics required to reproduce the motion.
\end{itemize}

\begin{figure*}[hbt]
    \centering
    \includegraphics[trim=0 10cm 0 0, clip,width=\linewidth]{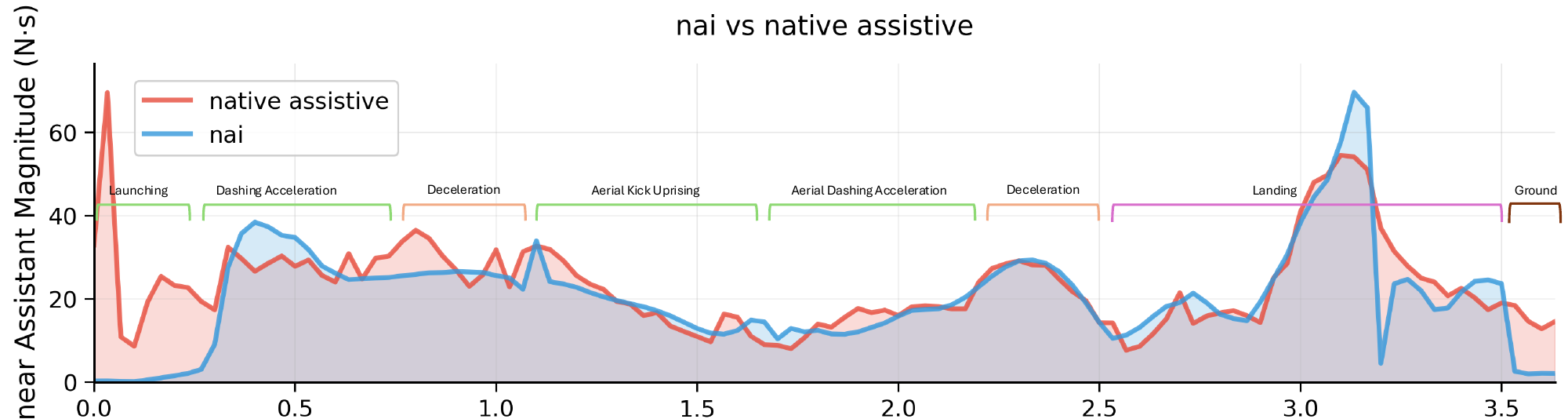}
    \caption{Comparison of assistive intervention profiles of teaser motion. The \textbf{Native Assist} (red) exhibits continuous, high-frequency force fluctuations ("always-on"), indicating an over-reliance on external assistance to force kinematic compliance. In contrast, \textbf{NAI} (blue) demonstrates distinct \textbf{sparsity}: the assistive impulse drops to near-zero during physically consistent phases (e.g., $t=0.0s-0.3s$, $t=3.5s+$) and activates smoothly only when dynamic transients require momentum injection.}
    \label{fig:impulse_profile}
\end{figure*}
\noindent{\textbf{Analysis of Results:} Table~\ref{tab:quantitative_results} reveals a critical trade-off between strict kinematic tracking and physical plausibility. 
While the naive method achieves a marginally lower tracking error ($E_{pose}$), this metric is \textbf{misleading}. 
The Baseline relies on excessive external impulses (Total $\mathbf{I}$) to brute-force the character into the reference pose, effectively ignoring the underlying physics. 
This results in ``over-fitted'' motion characterized by high-frequency oscillation and severe visual jittering (indicated by the high Jitter metric). 
Such artifacts render the motion visually jarring and physically unstable, despite the numerical closeness to the reference.

In contrast, \textbf{NAI} prioritizes physical integrity and temporal coherence. 
Although our tracking error is slightly higher than the over-fitted Baseline, it remains well within the standard range of state-of-the-art physics-based methods~\cite{2018-TOG-deepMimic, 2021-TOG-AMP,zhang2025ADD} at their regular motion tasks. 
This indicates that our method maintains high-fidelity tracking without resorting to continuous external intervention. 
Crucially, our Dual-Gated mechanism ensures significantly lower jitter, producing smooth, momentum-conserving motions. 
As visualized in Figure~\ref{fig:impulse_profile} (blue line), our controller exhibits a ``sparse'' activation pattern—remaining dormant during physically feasible phases and applying intervention only during necessary high-dynamic transients. 
This demonstrates that our higher smoothness and lower impulse cost represent a superior balance.

\subsection{Efficacy of Neural Residual Learning}
A core premise of our framework is that offline analytical solutions (RNEA) are insufficient for direct control due to the \textit{Sim-to-Data Gap}. Simulation introduces unpredictable dynamics, such as collision detection, friction variability, and discrete integration error, that an idealized rigid-body solver cannot foresee. Simply replaying an open-loop force trajectory results in temporal desynchronization, with the character's state lagging or leading the reference.

Figure~\ref{fig:impulse_profile} empirically validates the necessity of our learned residual term ($\mathbf{I}_{res}$). 
The orange line ($\mathbf{I}_{base}$) represents the ideal impulse calculated from the kinematic reference. While it captures the general trend of the motion, it frequently underestimates the required momentum magnitude needed in the actual simulation, particularly during high-contact-stress phases (e.g., $t=0.3\text{--}1.0s$ and $t=3.1\text{--}3.5s$).

Critically, the learned residual (green line, top) does not merely act as noise; it exhibits structured, meaningful intervention.
\begin{itemize}
    \item \textit{Magnitude Compensation:} When the analytical baseline cannot overcome simulation damping or contact loss, the residual branch generates a positive surge (e.g., the peak around $t=0.4s, 3.1s$) to "boost" the character, effectively closing the dynamics gap.
    \item \textit{Temporal Re-alignment:} The residual also adapts to timing mismatches. We observe phase shifts between the reference and the simulation (green), where the neural network dynamically modulates the impulse timing to match the character's instantaneous state ($s_t$) rather than the pre-recorded timeline.
\end{itemize}\vspace{-2mm}

This confirms that our Hybrid Architecture functions as intended: the analytical stream provides the macroscopic physical intent, while the neural stream acts as a closed-loop feedback controller, handling the complex, non-linear realities of the physics engine.

\subsection{Comparison with Pure Analytical Control (Open-Loop RNEA)}
\label{sec:pure_rnea}
\begin{figure}[hbt]
    \centering
    \includegraphics[width=0.99\linewidth]{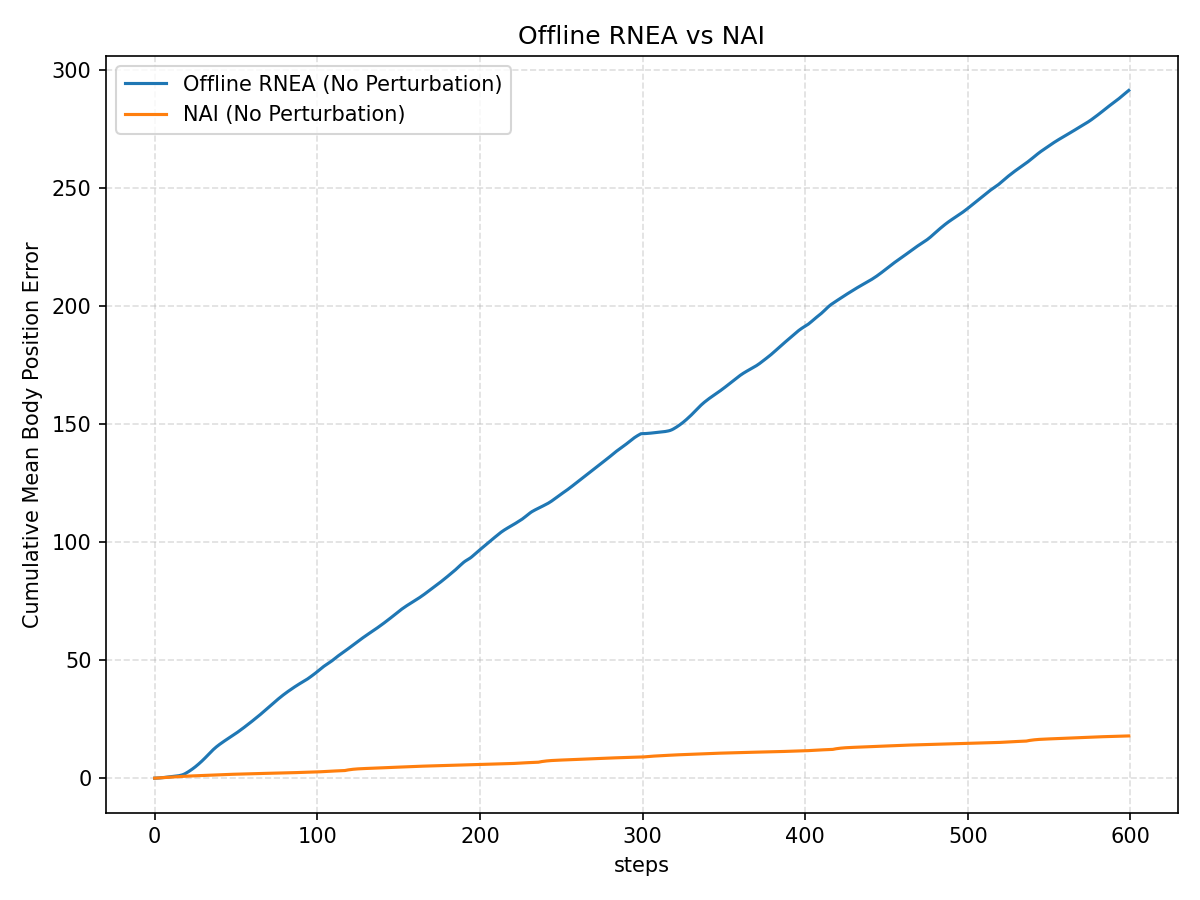}
    \caption{Tracking error comparison without external perturbations. The Open-Loop Offline RNEA (blue curve) accumulates numerical integration drift monotonically over time, leading to state divergence. The proposed closed-loop policy (orange curve) maintains bounded tracking errors.}

    \label{fig:rnea_error}
\end{figure}

To evaluate the mathematical necessity of the proposed closed-loop neural policy, we introduce a pure analytical baseline, denoted as Open-Loop RNEA. This experiment addresses the hypothesis of whether inverse dynamics alone is sufficient to track exaggerated motions within a discrete physics simulator.

\noindent{\textbf{Experimental Setup:}} For this comparative evaluation, the test is conducted specifically on a representative motion sequence—the continuous "Dashing Aerial Combat A" action featured in the teaser (see Fig. ~\ref{fig:teaser}). In this configuration, the neural residual policy is completely disabled. The character is actuated exclusively by the feedforward joint torques and the root assistive wrenches computed directly via the Recursive Newton-Euler Algorithm (RNEA) from the reference kinematics, which has been introduced in Sec. ~\ref{subsec:inv_dynamics}.. To test the dynamic robustness, we evaluate the system under two conditions: (1) standard tracking without external forces, and (2) a perturbation test wherein random external force vectors (e.g., projectile rigid body impacts) are applied to the character's rigid bodies during the execution. The experiments track the imitation target with 8 continuous episodes in 600 steps, with around 20 seconds.

\noindent{\textbf{Results and Analysis: }} The results demonstrate that the Open-Loop RNEA method fails to maintain long-term tracking stability with \textbf{0\% success rate}. As illustrated in Fig \ref{fig:rnea_error}, even in the absence of external perturbations, the open-loop character exhibits rapid and monotonic state drift, eventually triggering early termination conditions.
This tracking failure is caused by the inherent discrepancy between continuous analytical models and discrete numerical simulations. The RNEA formulates forces under the assumption of continuous-time dynamics and perfect state matching ($s_{sim} = s_{ref}$). However, physics engines (e.g., Isaac Gym) employ discrete numerical integration methods, such as the semi-implicit Euler method. Applying pre-computed analytical forces in a discrete environment inevitably introduces local truncation errors at each simulation time step $\Delta t$. 

Because the Open-Loop RNEA lacks a state feedback mechanism, these numerical integration errors accumulate monotonically over time, leading to an irreversible divergence between the simulated center-of-mass trajectory and the reference data. Furthermore, the pre-computed analytical forces contain no conditional logic regarding unpredictable environmental contacts. This result mathematically validates that integrating a closed-loop residual neural policy is essential; the policy functions as a dynamic feedback controller to proactively correct discrete integration drift and synthesize valid physical responses to external perturbations.
\subsection{Quantitative Analysis of Neural Residual Contribution}
\label{ssec:baseline_inaccuracy}

To further validate the necessity of the residual correction mechanism, we analyze the temporal activation of the gating scalars ($\beta_{lin}, \beta_{ang}$) during the execution of the ``Dashing Aerial Combat A'' sequence. This specific motion is characterized by highly dynamic, non-physical mid-air translations and instantaneous momentum shifts.

As illustrated in Figure \ref{fig:gate_values}, the neural policy dynamically modulates both the linear and angular gate values throughout the 120-step execution horizon. The empirical measurements indicate that both gating scalars fluctuate continuously within the $[0.3, 0.6]$ interval, converging to a mean activation of approximately $0.4$. A gating magnitude of $0.4$ signifies that the control policy allocates approximately $60\%$ of the assistive intervention to the learned neural residual ($\mathbf{I}_{res}$), while retaining only $40\%$ reliance on the pre-computed analytical baseline ($\mathbf{I}_{base}$).

\begin{figure}[htbp]
  \centering
  \includegraphics[width=0.99\linewidth]{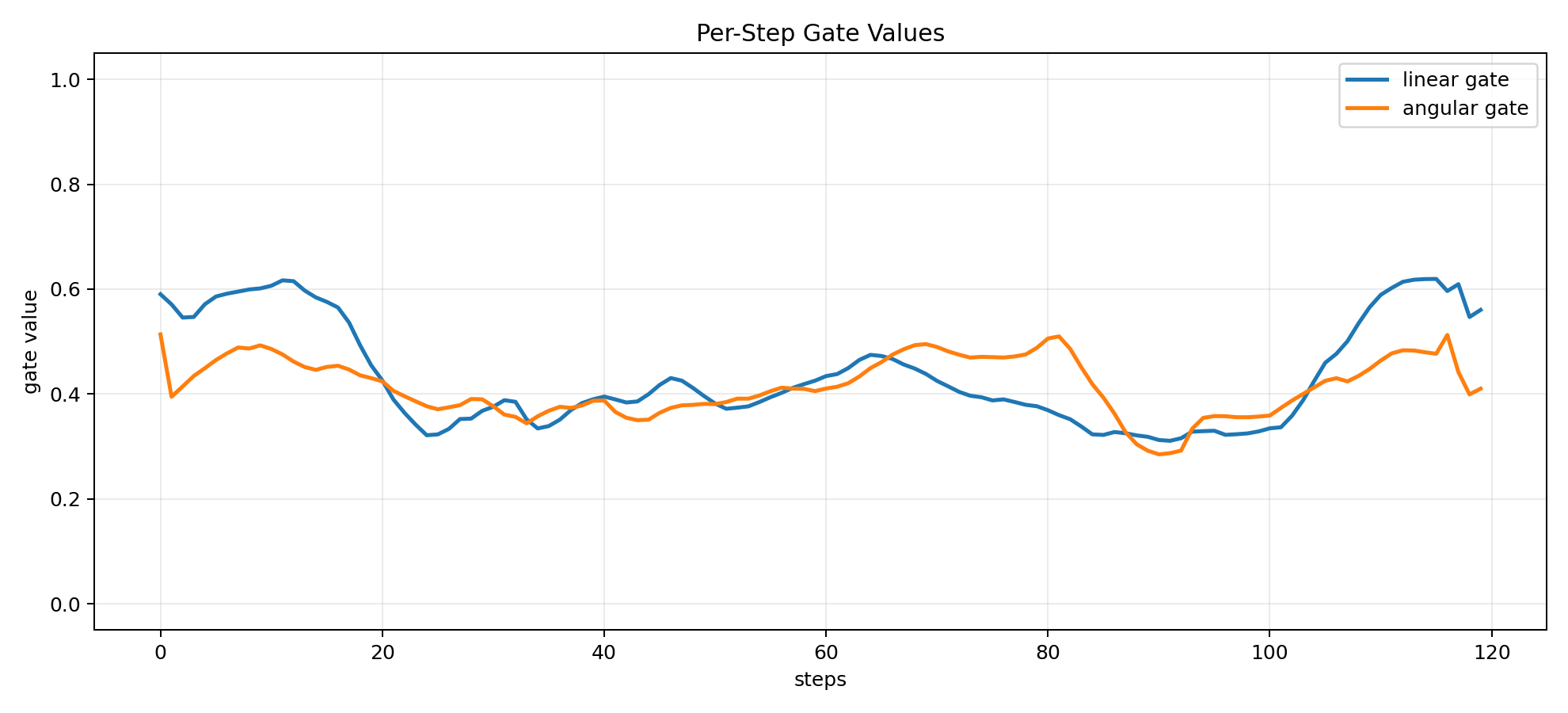}
  \caption{The linear ($\beta_{lin}$) and angular ($\beta_{ang}$) gating scalars. Both parameters are dynamically modulated within the $[0.3, 0.6]$ interval around a mean value of $0.4$. This indicates a consistent allocation of approximately $60\%$ of the assistive intervention to the neural residual.}
  \label{fig:gate_values}
\end{figure}

This persistent, non-zero residual activation quantitatively demonstrates the inherent numerical limitations of the offline Recursive Newton-Euler Algorithm (RNEA) when deployed in a forward simulation context. The continuous-time analytical formulation fails to perfectly map to the discrete integration steps ($\Delta t$) of the physics engine, particularly during periods of extreme kinematic acceleration. The empirical data confirms that the Neural Assistive Impulse (NAI) policy successfully identifies this dynamics gap, synthesizing the precise residual impulses required to correct the analytical approximation errors and enforce strict adherence to the target exaggerated trajectory.

\subsection{Dynamic Robustness Analysis}
\label{subsec:robustness}
\begin{figure}[hbt]
    \centering
    \includegraphics[width=0.99\linewidth]{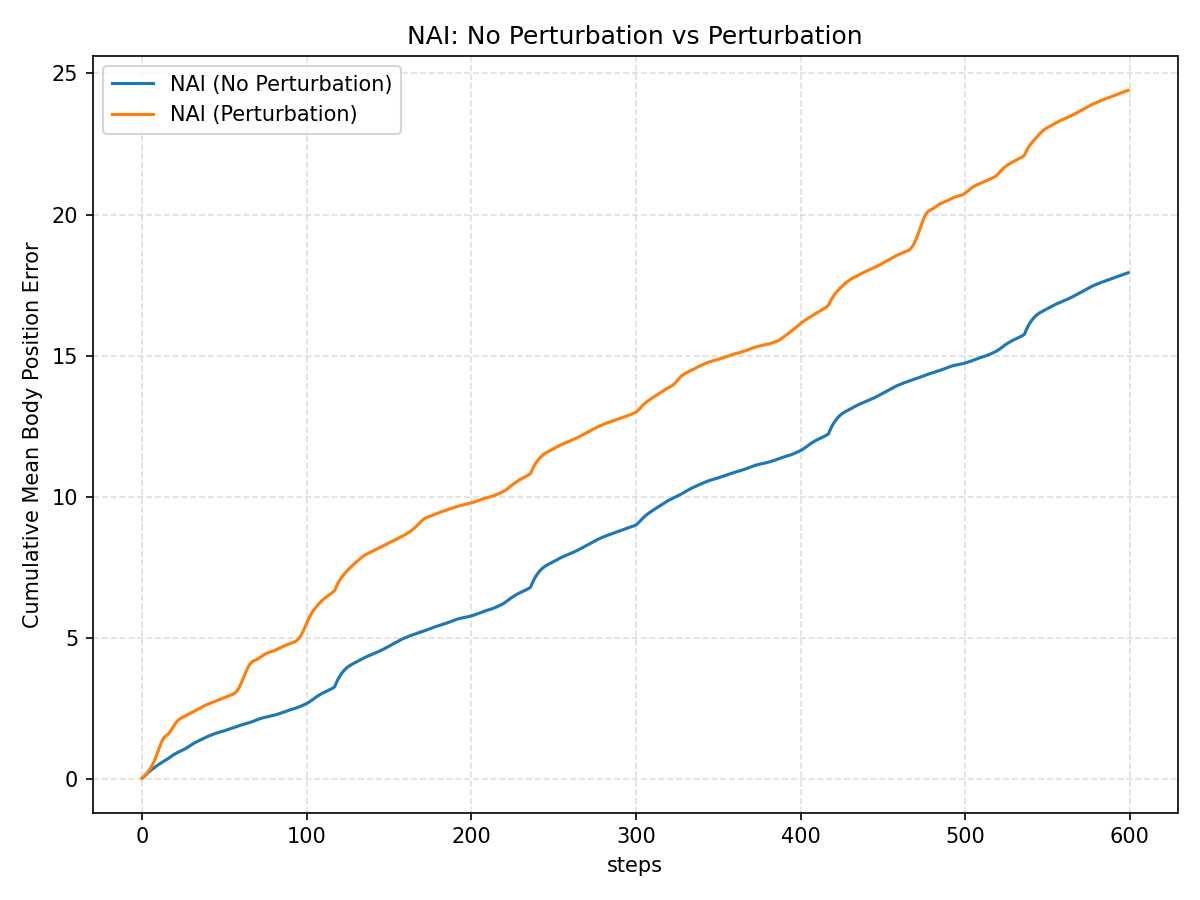}
    \caption{Tracking error comparison under external perturbation. Upon physical impact. The proposed closed-loop policy dampens the applied external impulse and synthesizes corrective actions to converge back to the reference trajectory.}

    \label{fig:rnea_error_turbulence}
\end{figure}
To quantitatively evaluate the dynamic robustness of the proposed Neural Assistive Impulse (NAI) framework, we conduct an interference analysis comparing the cumulative tracking error under standard and perturbed simulation conditions.

\noindent{\textbf{Experimental Setup:}} The system is evaluated under two distinct configurations over a continuous 600-step simulation horizon: (1) a baseline execution without external interference, and (2) a perturbed execution subjected to randomized physical turbulence. In the perturbed configuration, external spatial wrenches are injected into the character's torso at randomized intervals uniformly sampled between $20$ and $80$ simulation steps. Each perturbation event is sustained for a temporal window of $3$ to $12$ consecutive steps. The magnitudes of the applied linear forces and angular torques are uniformly sampled from the intervals $[100.0, 500.0]$ N and $[20.0, 50.0]$ N$\cdot$m, respectively. The primary metric is the Cumulative Mean Body Position Error, which quantifies the temporal accumulation of spatial deviation from the reference kinematics.

\noindent{\textbf{Results and Analysis:}} As illustrated in Figure \ref{fig:rnea_error_turbulence}, the proposed closed-loop NAI policy exhibits a bounded, non-divergent error accumulation rate in both configurations. The monotonic increase in cumulative error is a standard numerical property of discrete physical integration over extended horizons without global coordinate resets. 

While the perturbed scenario (orange curve) exhibits localized, step-wise displacements corresponding exactly to the instantaneous momentum injections from external impacts, the post-impact error derivative (slope) rapidly converges to match that of the unperturbed scenario (blue curve). This geometric parallelism demonstrates that the robust NAI policy, functioning as a state-conditioned reactive controller, immediately synthesizes corrective residual impulses ($\mathbf{I}_{res}$) to damp the perturbation and prevent trajectory divergence. Consequently, the system maintains a 100\% tracking success rate across 8 perturbed evaluation episodes, identical to its unperturbed baseline performance.
In contrast, as previously established in Section \ref{sec:pure_rnea}, an offline, open-loop control formulation (Pure RNEA) lacks this state-feedback mechanism. Under identical perturbation conditions, the open-loop system possesses zero capacity for dynamic error correction; its tracking error would accumulate unboundedly post-impact, leading to irreversible divergence and immediate simulation failure. The empirical ability of the NAI policy to maintain a controlled error derivative under severe external interference validates its mathematical necessity for robust physics-based motion synthesis.

\subsection{Ablation Study: Reward Formulation}
\label{ssec:ablation_loss}

To quantitatively isolate the contributions of the individual reward components, we conduct an ablation study focusing on the Shadow Compass Loss ($\mathcal{L}_{compass}$) and the Intervention Sparsity Loss ($\mathcal{L}_{sparsity}$). We evaluate the learning dynamics of the isolated models against the full Neural Assistive Impulse (NAI) framework by tracking the episode success rate and the mean body position error over 1000 training iterations. 

It is critical to note that the quantitative curves presented in Fig. \ref{fig:ablation_success_rate} and Fig. \ref{fig:ablation_body_error} reflect the performance during the continuous \textit{training} phase, rather than deterministic \textit{inference} evaluations. During training, the policy relies on stochastic action sampling for exploration and evaluates over continuous, concatenated motion loops. Consequently, the inherent randomness of the exploration policy and the accumulated difficulty of looping transitions yield lower absolute success rates and higher positional errors compared to the single-episode, deterministic evaluations executed during inference. Therefore, these curves primarily serve to illustrate the relative optimization efficiency and convergence stability among the different configurations.

\begin{figure}[ht]
  \centering
  \includegraphics[width=\linewidth]{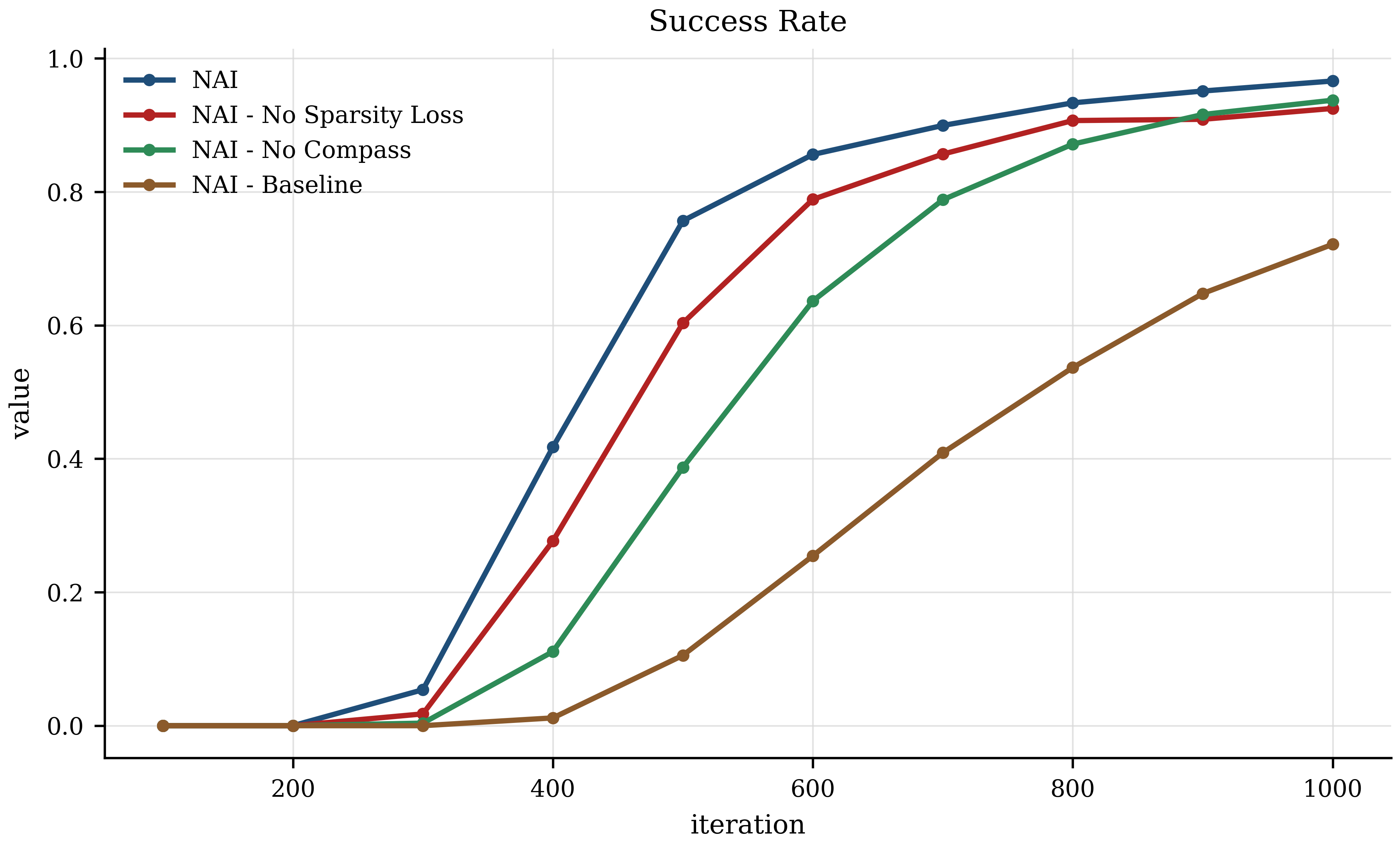}
  \caption{Ablation analysis evaluating the impact of the Shadow Compass and Sparsity loss formulations on the training success rate. The full NAI framework (blue curve) exhibits the most rapid convergence. Removing the Shadow Compass Loss (green curve) severely delays convergence due to inefficient exploration in the unguided directional space. The NAI - Baseline (brown curve), lacking both regularizations, exhibits the most severe optimization failure.}
  \label{fig:ablation_success_rate}
\end{figure}

\begin{figure}[ht]
  \centering
  \includegraphics[width=\linewidth]{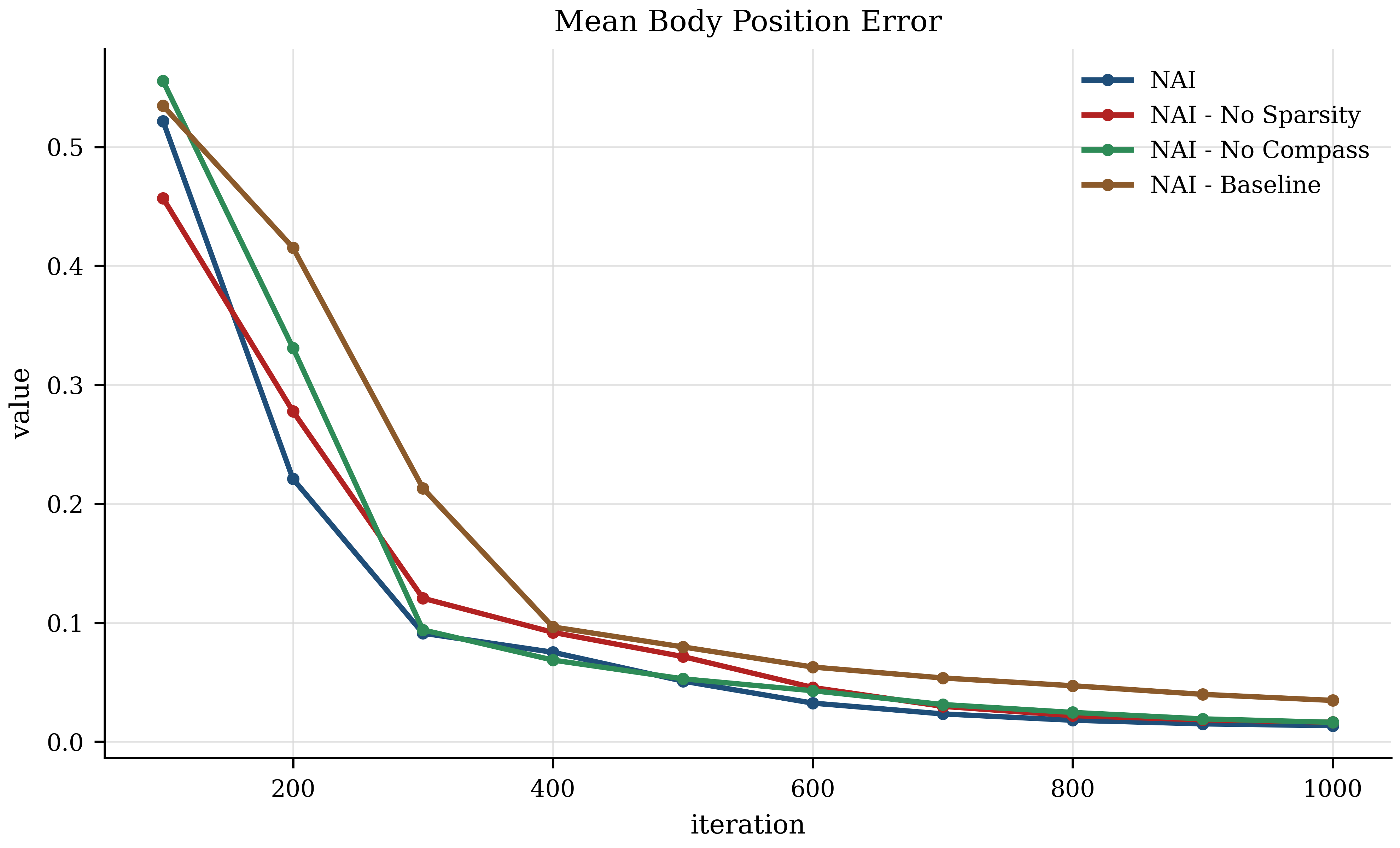}
  \caption{Ablation analysis evaluating the impact of the Shadow Compass and Sparsity loss formulations on the mean body position error. Removing the Sparsity Loss (red curve) introduces numerical drift from continuous unconstrained force injections, resulting in suboptimal error metrics during the mid-to-late training iterations. The NAI - Baseline (brown curve) consistently fails to minimize the tracking error effectively.}
  \label{fig:ablation_body_error}
\end{figure}
\noindent{\textbf{Effect of Combined Loss Removal (NAI - Baseline):}} 
The configuration omitting both regularization terms (denoted as NAI - Baseline) exhibits the most severe degradation in optimization efficiency. As illustrated by the brown curve in Figure \ref{fig:ablation_success_rate}, the success rate remains near zero for the initial 400 iterations and converges to a substantially lower terminal value compared to the regularized models. Correspondingly, the mean body position error (Figure \ref{fig:ablation_body_error}) remains persistently elevated throughout the training horizon. This configuration completely unconstrains the 6D wrench action space. In the absence of both geometric directional guidance and magnitude bounding, the policy relies exclusively on scalar tracking rewards. This induces an ill-conditioned credit assignment problem, causing the network to output unaligned and persistent momentum injections that exacerbate discrete numerical drift. The failure of this baseline empirically demonstrates the strict mathematical necessity of combining both constraints to successfully optimize the underactuated tracking problem within the standard iteration budget.

\noindent{\textbf{Effect of Shadow Compass Loss:}} 
The $\mathcal{L}_{compass}$ term is mathematically formulated to penalize the angular deviation between the synthesized residual impulse vector ($\mathbf{I}_{res}$) and the analytical kinematic trajectory derivative. As illustrated by the green curve in Figure \ref{fig:ablation_success_rate}, removing this directional constraint (denoted as NAI - No Compass) significantly decelerates the convergence speed of the success rate and causes the highest initial body position errors. Without explicit directional regularization, the optimization problem in the high-dimensional 6D wrench space becomes ill-posed, as the network relies solely on scalar positional tracking rewards. This lack of geometric guidance induces an inefficient credit assignment problem, delaying the policy's ability to synthesize assistive impulses that correctly align with the momentum requirements of the reference motion.

\noindent{\textbf{Effect of Intervention Sparsity Loss:}} 
The $\mathcal{L}_{sparsity}$ term functions as a regularization mechanism designed to minimize the magnitude of the continuous residual impulse ($m \to 0$) and maximize the reliance on the analytical baseline ($\beta \to 1$). The red curve (NAI - No Sparsity Loss) demonstrates that omitting this magnitude regularization results in suboptimal optimization efficiency. Specifically, between iterations 200 and 600, the model without sparsity regularization exhibits a slower reduction in mean body position error and a delayed success rate plateau compared to the full NAI framework. Without sparsity constraints, the neural network outputs persistent, unconstrained residual forces regardless of the underlying physical necessity. This continuous injection of non-physical momentum overrides the rigid-body dynamics solver, introducing discrete numerical drift during the continuous motion loops of the training phase. By enforcing $\mathcal{L}_{sparsity}$, the framework mathematically bounds the residual interventions, minimizing unnecessary numerical accumulation and accelerating stable convergence.

\noindent{\textbf{Quantitative Analysis of Control Jitter:}} 
To further evaluate the numerical stability of the control policies, we analyze the control signal jitter across the ablation configurations. The empirical jitter metrics are summarized in Table \ref{tab:ablation_jitter}. 

\begin{table}[htbp]
  \centering
  \caption{Quantitative comparison of control signal jitter across ablation configurations. The Sparsity formulation eliminates the oscillations of the baseline while preserving the necessary high-frequency impulse spikes (yielding a nominally higher jitter than the over-damped, non-physical No-Sparsity configuration).}
  \label{tab:ablation_jitter}
  \begin{tabular}{lc}
    \toprule
    \textbf{Configuration} & \textbf{Control Jitter} \\
    \midrule
    NAI - Baseline & 3.45 \\
    NAI - No Sparsity & 2.14 \\
    NAI - No Compass & 2.04 \\
    \textbf{Full NAI (Ours)} & \textbf{2.29} \\
    \bottomrule
  \end{tabular}
\end{table}

The unconstrained NAI-Baseline exhibits the highest jitter magnitude ($3.45$). In the absence of magnitude and directional regularizations, the policy outputs erratic, high-frequency oscillatory wrenches, causing severe numerical instability in the dynamics solver. By applying the combined loss formulation, this pathological noise is significantly suppressed. Notably, the full NAI policy exhibits a marginally higher nominal jitter ($2.29$) compared to the unregularized NAI-NoSparsity configuration ($2.14$). From a continuous dynamics perspective, this is a mathematically rigorous outcome. The $\mathcal{L}_{sparsity}$ constraint forces the assistive interventions to remain strictly at zero during physically valid segments and to activate exclusively as sharp, discrete impulse spikes at kinematic discontinuities. These necessary instantaneous momentum injections inherently register as localized high-frequency components in the derivative of acceleration (jitter). In contrast, the NAI-NoSparsity configuration artificially lowers the overall jitter metric through the continuous temporal dispersion of assistive forces—an over-damped, non-physical behavior that overrides the underlying rigid-body dynamics and ultimately degrades the tracking success rate. Thus, the Sparsity formulation successfully eliminates the pathological numerical oscillations of the baseline while preserving the sharp, high-frequency impulse spikes mathematically essential for executing exaggerated maneuvers.

\section{Conclusions \& Limitations}\label{sec:conclusion}
We have presented a novel framework for physically simulating exaggerated, stylized character motions—a domain that has traditionally been intractable for standard physics-based controllers.

By fundamentally shifting the control method from Force Space to Momentum Space, we resolve the numerical instabilities inherent in tracking high-frequency, physically infeasible maneuvers. Our core contribution, the \textit{Hybrid Dynamics Architecture}, successfully bridges the gap between kinematic imagination and dynamic reality. By synergizing an open-loop analytical baseline (derived from RNEA) with a closed-loop neural residual, our system enables characters to execute "physics-defying" skills—such as mid-air dashes and instantaneous trajectory changes—while maintaining robustness and minimizing "ghost force" artifacts.

First, the quality of the analytical baseline ($\mathbf{I}_{base}$) is heavily dependent on the kinematic consistency of the reference motion. If the source animation contains severe interpenetrations or non-smooth noise, the RNEA solver may produce erratic guidance, forcing the neural residual to overcompensate. Second, while our \textit{Shadow Compass} and sparsity objectives effectively regularize the assistance, tuning the trade-off between strict kinematic tracking and physical plausibility remains a task-dependent process. 
Future work anticipates integrating this framework with \textit{Generative AI} pipelines. Current text-to-motion diffusion models frequently produce imaginative yet physically invalid animations. Our Momentum-Space control could serve as a robust “physics adapter,” anchoring these generative hallucinations within interactive, simulated environments. Additionally, extending our residual formulation to accommodate object interactions (e.g., stylized weapon combat with destructible environments) presents a promising direction for future research.

% bibtex
\bibliographystyle{eg-alpha-doi} 
\bibliography{references}

\end{document}